\documentclass[conference]{IEEEtran}
\usepackage{newtxtext}
\IEEEoverridecommandlockouts
\usepackage{cite}
\usepackage{amsmath,amssymb,amsfonts}
\usepackage{algorithmic}
\usepackage{graphicx}
\usepackage{textcomp}
\usepackage{xcolor}
\usepackage{balance}
\def\BibTeX{{\rm B\kern-.05em{\sc i\kern-.025em b}\kern-.08em
    T\kern-.1667em\lower.7ex\hbox{E}\kern-.125emX}}
\begin{document}

\title{A Rapid Pipeline for Training and Deploying ML Models on WeBe Band
}

\author{\IEEEauthorblockN{1\textsuperscript{st} Ehsan Kourkchi$^*$ \thanks{* a.k.a Edwin E. Kay (ekay@ucdavis.edu)}}
\IEEEauthorblockA{
\textit{Dep. of Electrical and Computer Engineering} \\
UC Davis, CA, USA. \\
ORCID: 0000-0002-5514-3354}
\and
\IEEEauthorblockN{2\textsuperscript{nd} Asmita Asmita}
\IEEEauthorblockA{
\textit{Dep. of Electrical and Computer Engineering} \\
UC Davis, CA, USA. \\
aasmita@ucdavis.edu}
\and
\IEEEauthorblockN{3\textsuperscript{rd} Houman Homayoun}
\IEEEauthorblockA{
\textit{Dep. of Electrical and Computer Engineering} \\
UC Davis, CA, USA. \\
hhomayoun@ucdavis.edu}
\and
\IEEEauthorblockN{4\textsuperscript{th} Mahdi Eslamimehr}
\IEEEauthorblockA{\textit{dept. name of organization (of Aff.)} \\
\textit{Quandary Peak Research} \\
Los Angeles, CA, USA. \\
mahdi@quandarypeak.com}
}

\maketitle




\begin{abstract}

Developing optimized machine-learning algorithms for edge devices with limited computational and memory resources is challenging, time-consuming, and highly dependent on device-specific constraints. In this work, we streamline an edge ML workflow to enable rapid development, optimization, and deployment of machine-learning (ML) models directly on the \emph{WeBe Band}, a wrist-worn wearable device designed for multimodal physiological data monitoring. The proposed system automatically generates hardware-efficient ML models that can be easily integrated into the WeBe core firmware, supporting AutoML, hardware-aware quantization, and performance profiling to build models that meet desired latency targets while remaining compatible with device memory and power limitations. 

The proposed framework tightly integrates the open source {\it Piccolo AI} ecosystem with an automated pipeline that generates deployable firmware artifacts, performs hardware-aware model compilation, and supports over-the-air (OTA) deployment. The system supports multiple lightweight model classes, including classical machine-learning algorithms and neural networks, and provides built-in on-device profiling tools to evaluate inference latency and memory footprint under realistic execution conditions. Experimental results demonstrate clear trade-offs between model complexity and deployability on a microcontroller, showing that classical models offer strong real-time performance while lightweight neural networks require careful resource management.

Rather than proposing new learning architectures, the current work mainly focuses on system-level automation, deployability, and enabling researchers and developers to rapidly iterate on models and evaluate them directly on target hardware. Although demonstrated on the \emph{WeBe Band} platform, the workflow is designed to be extensible to other ML-powered edge devices.

\end{abstract}


\section{Introduction}

Tiny Machine Learning (TinyML) has become a powerful way to run machine learning (ML) models directly on small, low-power edge devices such as microcontrollers, wearable devices, and IoT systems \cite{banbury2021benchmarkingtinymlsystemschallenges, warden2019tinyml, Dutta2021TinyMLMI}. It enables real-time inference on ultra-low-power hardware, reduces system latency, and minimizes reliance on cloud computing. TinyML is now widely used in many applications, including wearable health monitoring \cite{SOUMMA2025100628}, industrial machinery maintenance and anomaly detection \cite{industrial}, and environmental or smart-home automation \cite{Ken2025}. Despite its growing use in consumer and industrial products, the adoption of tiny ML models in academic and applied research remains relatively limited. One major reason is the fragmented workflow between model development and deployment on embedded hardware. Researchers are often expected to have expertise not only in machine learning, but also in embedded systems, including firmware development, compiler toolchains, and hardware-level optimization. This steep technical barrier can slow progress, particularly for domain experts in areas such as healthcare or physiological signal analysis, whose primary focus is on data interpretation rather than low-level system integration.



In wearable health research, this gap between model development and embedded deployment is particularly pronounced. Many interdisciplinary teams possess strong expertise in data science or clinical analysis but lack the embedded systems knowledge required to transform machine-learning models into real-time applications on wearable hardware. As a result, promising ML models for edge devices frequently remain confined to offline evaluation environments rather than being tested under realistic sensing and execution conditions. Bridging this gap requires workflows that not only optimize model performance but also streamline integration, compilation, and deployment directly on target devices.



In this work, we present a research framework to abstract away the complexities of deploying optimized ML models on embedded devices. Our proposed system provides an end-to-end pipeline for \emph{WeBe Band} that enables researchers to experiment with diverse ML models and feature sets without manually interacting with the device firmware. The workflow allows users to simply upload collected data and the corresponding ML model artifacts; and our backend framework automatically generates the appropriate header files, configures compiler settings, and produces a deployable binary or ZIP package. This package can then be flashed or delivered via over-the-air (OTA) to \emph{WeBe Band}. By automating the integration process, the proposed pipeline allows researchers to mainly focus on model development, performance evaluation, and domain-specific analysis rather than low-level firmware handling. We anticipate that this system will lower the entry barrier for TinyML experimentation and accelerate research in edge-based healthcare analytics.

The framework supports multiple lightweight model classes, including classical machine-learning algorithms and neural networks, and provides built-in profiling tools that measure inference latency and memory footprint directly on-device. This hardware-aware evaluation allows users to explore the trade-offs between model complexity and real-time feasibility under realistic microcontroller constraints. 


Although demonstrated on the \emph{WeBe Band} platform, the architectural design of the workflow is intended to be extensible to other Cortex-M class devices, highlighting a broader shift toward deployability-focused TinyML research. Rather than proposing new learning architectures, this work emphasizes the importance of integrated development pipelines that enable rapid experimentation and practical deployment in healthcare-oriented edge AI systems. The resulting framework provides a scalable foundation for interdisciplinary research, allowing developers, clinicians, and data scientists to collaboratively explore edge ML applications directly on wearable devices.

The main contributions of this paper are summarized as follows:
\begin{itemize}
    \item We present an end-to-end TinyML workflow that automates model integration, firmware compilation, and OTA deployment on a wearable platform.
    \item We provide direct on-device profiling of latency and memory footprint across multiple model architectures, highlighting system-level trade-offs under realistic constraints.
    \item We demonstrate how automated deployment pipelines can accelerate rapid prototyping and hardware-aware model selection for embedded applications.
\end{itemize}

\begin{figure}[t]
\centerline{\includegraphics[width=\columnwidth]{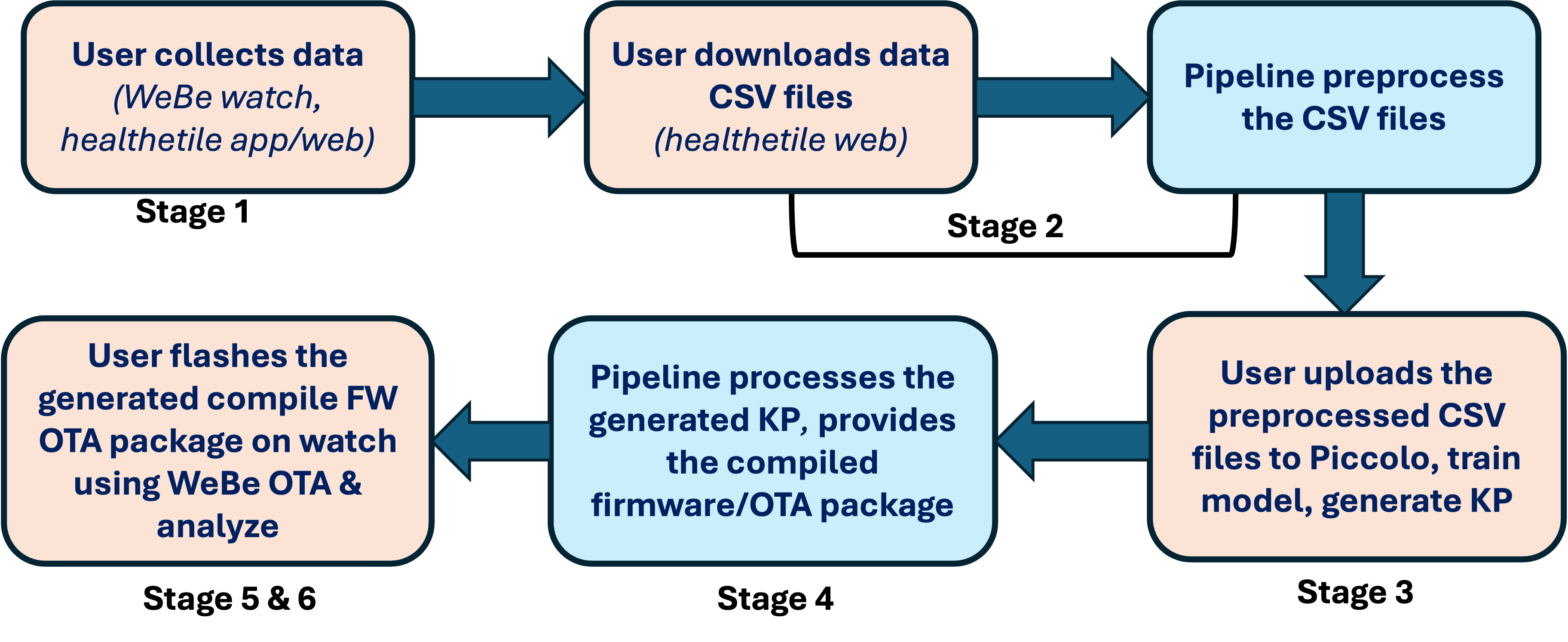}}
\caption{Overview of the proposed end-to-end pipeline. KP stands for the model knowledge pack, a ZIP file that holds the model artifacts including the compiled flashable library.}
\label{pipeline}
\end{figure}
\section{Background}

\subsection{Related Work}

Existing edge AI frameworks and toolchains have primarily focused on optimizing inference performance and model compression, rather than simplifying the end-to-end workflow for research deployment. Banbury et al. \cite{banbury2021benchmarkingtinymlsystemschallenges} and Warden et al. \cite{warden2019tinyml} discuss the challenges of bringing ML to microcontrollers, highlighting constraints in memory and computation. However, their solutions are mostly centered on benchmarking and runtime optimization.

Several studies have proposed TinyML-based applications targeting specific domains such as industrial IoT \cite{industrial} and environmental monitoring \cite{Ken2025}, demonstrating the potential of on-device intelligence. However, these works often assume pre-existing embedded expertise and require manual firmware adaptation. For health-oriented applications, \cite{SOUMMA2025100628} discusses ML-based analytics for wearable health monitoring. Although these approaches showcase promising ML models for clinical decision making, they lack a generalized workflow that abstracts device-specific hardware configurations. Although prior TinyML frameworks provide efficient runtime environments or optimized inference kernels, the integration of trained models into production-ready firmware remains largely manual and device-specific. Researchers often need to modify embedded codebases, adjust memory layouts, and handle hardware-dependent deployment steps, creating a significant barrier for interdisciplinary teams working in healthcare and wearable sensing. Consequently, there exists a gap between model-centric embedded model research and practical deployment workflows. 

\subsection{WeBe Band}

\emph{WeBe Band}, developed by Health-eTile, is a research-grade wearable device designed for high-fidelity physiological and activity monitoring \cite{webe_pubmed, webe_band_healthetile, healthetile_website}. The device integrates an array of sensors, including four photoplethysmography (PPG) channels (two Green channels, Red, and Infrared), a tri-axial accelerometer, skin temperature sensor, and electrodermal activity (EDA), which reflects skin conductance linked to sympathetic nervous system activity. The WeBe platform also provides physiological and contextual indicators, including estimated heart rate, oxygen saturation (SpO$_2$), skin contact state, walking step count, and actigraphy metrics. Together, these capabilities make \emph{WeBe Band} a promising platform for ambulatory medical monitoring, rehabilitation tracking, elderly care, and real-time patient supervision in both hospital and home settings. The WeBe ecosystem is supported by both a mobile application and a web portal interface, which enable users and researchers to collect, visualize, and manage physiological data streams in real time. In addition, the platform provides a Python SDK for data access, analysis, and offline processing, as well as mobile API supporting programmatic device control, data synchronization, and customizable application-level interactivity. These software components expose structured interfaces for streaming, querying, and managing multimodal physiological data, facilitating integration with external analytics pipelines, cloud services, and research workflows. 

Given its multimodal sensing capabilities and extensible software ecosystem, \emph{WeBe Band} serves as an ideal platform for evaluating and deploying tiny healthcare models on edge devices. The device supports configurable sampling rates, real-time Bluetooth streaming, and secure offline data storage, enabling both continuous ambulatory monitoring and controlled experimental studies. Integrated over-the-air (OTA) firmware update capability further allows rapid deployment of algorithmic improvements and system-level modifications without physical device access. In addition to raw sensor acquisition, the platform supports on-device preprocessing, feature extraction, and embedded inference, facilitating hardware-aware evaluation of machine-learning models under realistic operating conditions. \emph{WeBe Band} is built around an ARM Cortex-M4F microcontroller ({\it nRF52840}, 64 MHz) featuring 1 MB of on-chip RAM, and incorporates 256/512/1024/2048 Mbits of external flash memory to support extended data logging, firmware management, and edge-model deployment \cite{webe_pubmed, fang2024webe}. The platform supports multiple configurable sampling rates (e.g. 25, 50, 64, 86, and 100 Hz) enabling adaptation to diverse physiological monitoring and activity-recognition scenarios.

\subsection{Piccolo AI}

{\it Piccolo AI} is an open-source, lightweight, edge-focused machine learning framework developed by {\it SensiML} that enables researchers and developers to efficiently create, train, and deploy tiny models on embedded devices \cite{piccolo_github, piccolo_blog}. The platform provides a comprehensive interface through which users can explore a variety of feature types, including statistical, temporal, and spectral features. {\it Piccolo AI} supports several standard machine-learning model types, such as Random Forest, Bonsai, and Pattern Matching Engine (PME), as well as neural network architectures (fully connected, convolutional, and temporal), and offers a flexible environment for training, optimization, and model tuning according to application-specific requirements. Upon completion of training, the framework allows users to export the generated model as a knowledge pack (KP), i.e., a ZIP archive containing all necessary artifacts for deployment.

In our research workflow, {\it Piccolo AI} serves as the primary environment for model development. Users can train models with different datasets and feature configurations, and then provide the resulting knowledge pack to our pipeline to prepare it for deployment on \emph{WeBe Band}. Our framework automates all background steps, including generating the required embedded header files, configuring compilation settings, and producing a fully deployable binary package. This compiled ZIP can then be flashed onto \emph{WeBe Band} using the Bluetooth over-the-air (OTA) system, without requiring users to deal with the internal structure of the hardware or firmware. By integrating {\it Piccolo AI} with our pipeline, researchers are able to experiment with various ML models, explore numerous feature combinations and training configurations, and subsequently evaluate model performance on the desired edge device. This approach effectively abstracts the complexity of embedded deployment, enabling rapid prototyping and experimentation on a wearable health platform.

Although the current implementation targets the ARM Cortex-M4F microcontroller used in \emph{WeBe Band}, and the overall pipeline design is not inherently device-specific. The automated workflow relies on model knowledge packs, header generation, and firmware abstraction layers that can be adapted to other Cortex-M class platforms such as STM32, ESP32, or Nordic nRF53-series devices with minimal architectural changes. Consequently, the presented system should be viewed as a generalizable workflow framework, with \emph{WeBe Band} serving as a reference implementation for wearable edge AI deployment.

\begin{table*}[t]
\centering
\caption{Profiling metrics of the ML models evaluated on \emph{WeBe Band}.}
\label{tab:generic-metrics}

\setlength{\tabcolsep}{5pt}  
\small                      

\begin{tabular}{c|c|c|c|c|c}
\hline
(1) & (2) & (3) & (4) & (5) & (6) \\
\textbf{Model*} & \textbf{Architecture} & \textbf{\# Params} & \textbf{Latency ($\mu$s)} & \textbf{Flash (Bytes)} & \textbf{SRAM (Bytes)} \\ 
\hline \hline
RF  & -- & -- & 8520 & 10368 & 1244 \\ \hline
PME & -- & -- & 8763 & 8234  & 1364 \\ \hline
NN1 & 32--32--16--8  & $\sim$3.9K & 10535 & 15666 & 4444 \\ \hline
NN2 & 64--64--32--16--8  & $\sim$8.9K & 11209 & 24050 & 4444 \\ \hline
NN3 & 64--32--16--8  & $\sim$5.9K & 10769 & 18930 & 4444 \\ \hline
NN4 & 32--32--16--16--8  & $\sim$4.8K & 10832 & 16754 & 4444 \\ \hline

\multicolumn{6}{l}{} \\

\multicolumn{6}{l}{\footnotesize *~PME = Pattern Matching Engine; RF = Random Forest; NN = Neural Network.} \\
\multicolumn{6}{l}{\footnotesize Column (1) lists the evaluated model.} \\
\multicolumn{6}{l}{\footnotesize Column (2) denotes the hidden-layer configuration.} \\
\multicolumn{6}{l}{\footnotesize Column (3) indicates the approximate number of trainable parameters.} \\
\multicolumn{6}{l}{\footnotesize Column (4) reports maximum on-device inference latency ($\mu$s).} \\
\multicolumn{6}{p{0.6\textwidth}}{\footnotesize Columns (5) and (6) represent the Flash memory required for model storage and the SRAM required during runtime execution.} \\
\end{tabular}
\end{table*}
\section{Design and Implementation}

The general workflow of our system, illustrated in Fig.~\ref{pipeline}, consists of six primary stages that enable end-to-end automation from data acquisition to model deployment and analysis.

\subsubsection{Stage 1: Data Collection}

In the initial phase, physiological and motion data are collected at a sampling rate of 25~Hz using \emph{WeBe Band} in conjunction with its companion mobile and web applications~\cite{webe_band_healthetile, healthetile_website}. For the purpose of evaluating the performance of different models with varying complexity on the WeBe processor, this study focuses exclusively on motion detection algorithms. We collected data for six different gestures performed by two users, consisting of drawing the six letters {\it A}, {\it B}, {\it C}, {\it D}, {\it X}, and {\it O} in space in a consistent manner. The amount of data collected and the absolute precision of the model are not the primary focus of this study; instead, we emphasize the model-building process, the deployment, and the profile of the models generated on \emph{WeBe Band}.

\subsubsection{Stage 2: Data Pre-processing}

The preprocessing stage follows our automated pipeline standards to prepare the collected CSV data for downstream tasks. This process includes handling missing values, synchronizing timestamps, and normalizing sensor measurements. The data fields are also formatted to meet the firmware requirements of the WeBe device. All gestures were segmented into windows of $\gtrsim$100 samples (approximately 4 seconds) and labeled manually or semi-automatically using the {\it Piccolo AI} labeling interface.

\begin{figure}[t]
\label{fig:umap_accuracy}
\centering
\begin{minipage}[t]{0.4\textwidth}
    \centering
    \includegraphics[width=\linewidth]{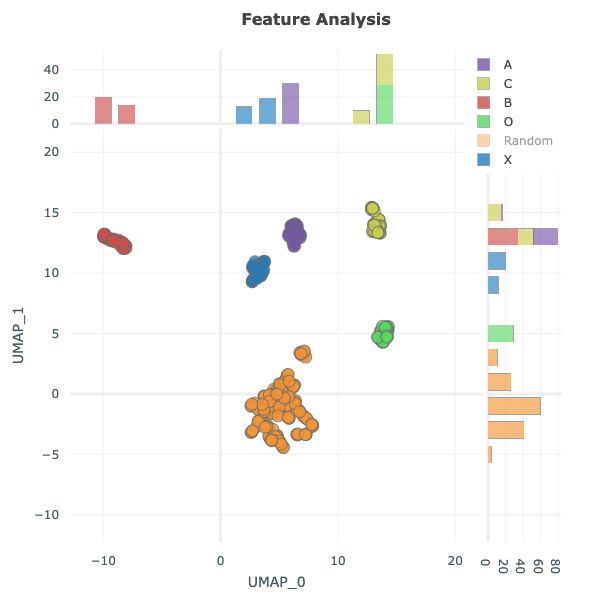}
\end{minipage}
\hfill
\begin{minipage}[t]{0.5\textwidth}
    \centering
    \includegraphics[width=0.9\linewidth]{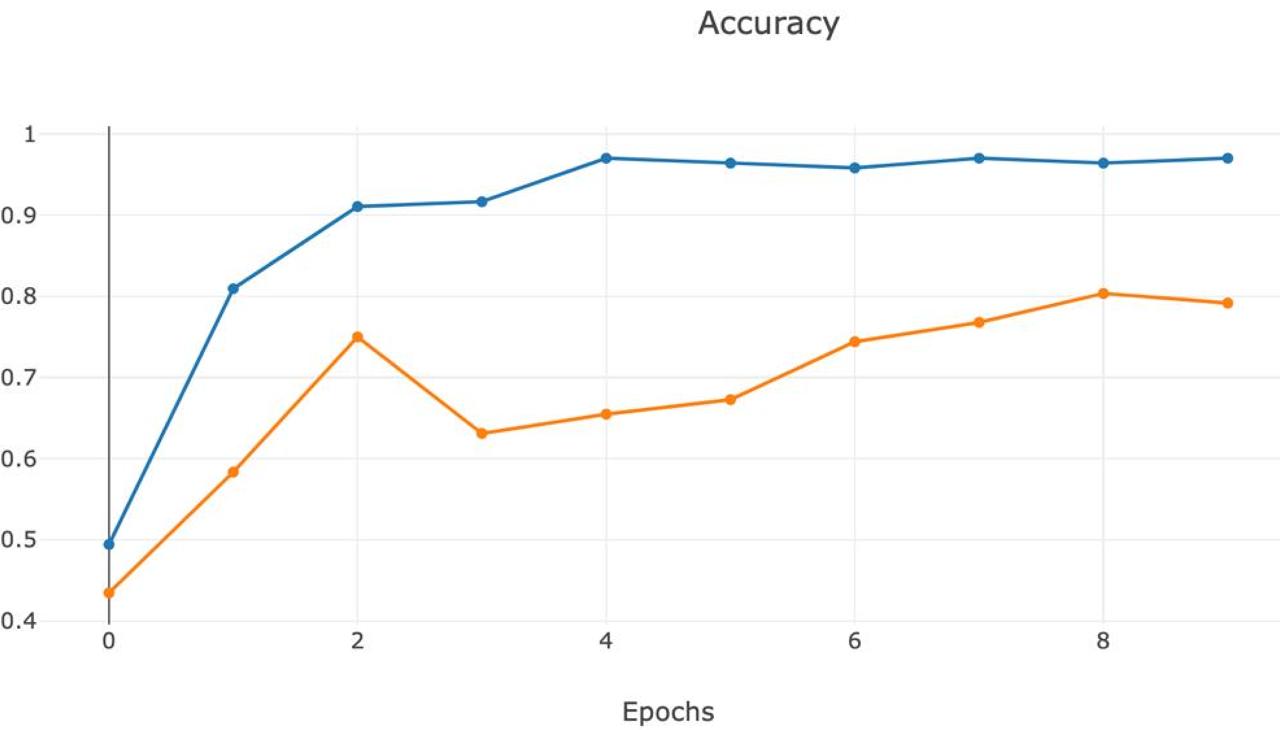}
\end{minipage}
\caption{{\bf Top:} Visualization of the 66 extracted features using the UMAP dimensionality-reduction method. The clustering structure illustrates how the {\it Piccolo AI} framework identifies discriminative feature subsets that effectively separate multiple gesture labels in feature space. {\bf Bottom:} Training and validation accuracy as a function of epoch for the neural network architecture 64--64--32--16--8 (see Table~\ref{tab:generic-metrics}). The blue curve represents validation accuracy, while the orange curve corresponds to training accuracy. Batch normalization and dropout regularization are applied during training, which explains the slight gap between the curves and helps mitigate overfitting.}
\label{fig:two_panel}
\end{figure}

\subsubsection{Stage 3: Model Generation}

For model generation, we utilize the {\it Piccolo AI} framework~\cite{piccolo_github, piccolo_blog}. Users import the preprocessed CSV files and configure the feature extraction queries. The {\it Piccolo AI} environment allows for flexible automatic or manual selection and tuning of various types of features, such as statistical, temporal, and spectral. Upon completion of the training process, the final models are compiled and downloaded as a Cortex-M compatible Knowledge Pack (KP), which encapsulates all artifacts and the model inference pipeline in a ZIP package ready for deployment on compatible edge devices.

In this study, we explored three machine-learning algorithms, as described below.

\begin{itemize}

\item \textbf{Pattern Matching Engine (PME):} 
PME is a distance-based classifier that compares incoming feature vectors against a set of stored prototypes using metrics such as $L_1$ distance \cite{hammer2011prototype}. Each prototype is associated with an \emph{Area of Influence (AIF)}, which defines the similarity boundary for classification. In our experiments, the AIF range was constrained between 25 and 400 to balance sensitivity and robustness while preventing excessive model growth. The model dynamically allocates up to 128 neurons to adequately cover the feature space while maintaining a compact memory footprint suitable for embedded deployment.

\item \textbf{Random Forest (RF):} 
Random Forest is a classical ensemble-learning method that aggregates predictions from multiple decision trees using majority voting. Its robustness to noisy sensor inputs and resistance to overfitting make it well suited for motion-recognition tasks on wearable platforms. We trained a forest consisting of 40 decision trees with a maximum depth of 7, selected to achieve a balance between classification performance and computational efficiency under microcontroller constraints.

\item \textbf{Neural Network (NN):} 
We evaluated lightweight, fully connected neural networks to investigate the trade-off between model expressiveness and resource usage. Several architectures were explored, as listed in Table \ref{tab:generic-metrics}. Models were trained using the Adam optimizer with a learning rate of 0.0015, a batch size of 32, and a dropout rate of 0.1 to mitigate overfitting. Batch normalization was applied to improve training stability. During deployment, model outputs were post-processed using a conservative 60\% confidence threshold, mapping low-confidence predictions to an ``Unknown'' class to reduce false positives in real-time operation. 
\end{itemize}

It is worth noting that classification accuracy and generalization performance are not the primary focus of this study; instead, the objective is to evaluate deployability, latency, and memory behavior under realistic embedded constraints. The UMAP feature visualization and sample training curves are presented in Fig.~\ref{fig:two_panel}, illustrating the stability of the training process and the discriminative structure of the feature set.

\subsubsection{Stage 4: Model Integration and Firmware Compilation}

Once a user provides the generated model knowledge pack (KP), our automated backend framework orchestrates its seamless integration into the WeBe firmware. This involves parsing the KP, dynamically generating the necessary header files, embedding them into the firmware source tree, and compiling a complete, deployable binary. The framework abstracts the underlying complexity of firmware modification and compilation by delivering a pre-compiled firmware and/or over-the-air (OTA) update package without requiring users to interact with the embedded toolchain.

\subsubsection{Stage 5: Firmware Flashing}

The compiled OTA package can be directly flashed onto the WeBe device through Bluetooth data packets using a single command provided within our pipeline interface. This OTA update mechanism significantly simplifies deployment by eliminating the need for physical device interface or manual firmware flashing procedures.

\subsubsection{Stage 6: On-Device Evaluation and Analysis}

After deployment, users can execute profiling and performance analysis scripts, either custom or provided by our framework, to evaluate the behavior of the models on live sensor data. Metrics such as inference latency can be monitored and compared across models. This streamlined loop allows users to iteratively experiment with different ML models and optimize their hyperparameter and feature sets, as they can be rapidly prepared for live execution on the WeBe device.

\subsection{Hardware Portability Considerations}

Although the present study demonstrates the pipeline on \emph{WeBe Band}, the workflow is designed around modular firmware integration principles to facilitate portability across embedded platforms. The model knowledge pack exported from {\it Piccolo AI} is parsed independently of the underlying hardware, and allows header files and inference pipelines to be generated dynamically for different firmware targets. 
Similarly, the OTA packaging process abstracts device-specific flashing procedures, and can be easily adapted to alternative embedded hardware with comparable memory and compute characteristics. Future efforts involve exploring broader compatibility with heterogeneous microcontroller ecosystems and further validating the applicability of the proposed framework.

\begin{figure*}[htbp]
\centerline{\includegraphics[width=\linewidth]{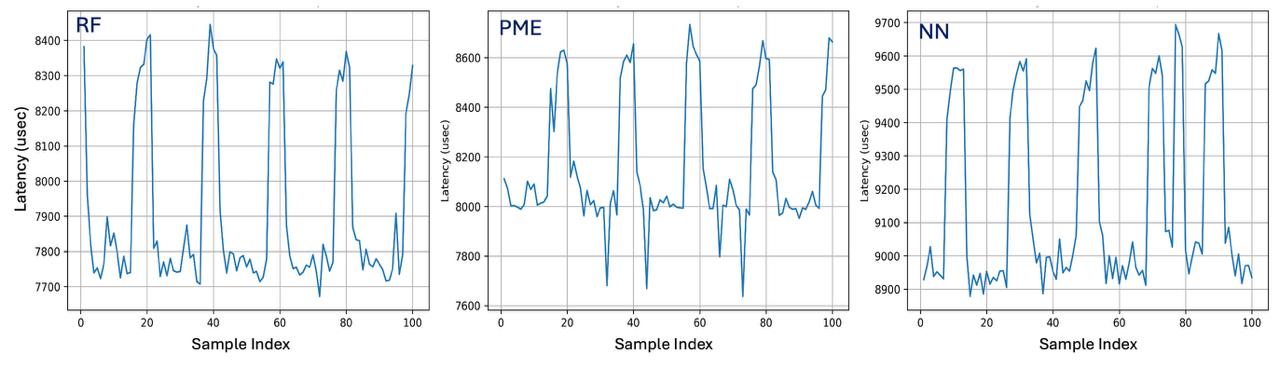}}
\caption{Latency versus sample index for Random Forest (RF), Pattern Matching Engine (PME), and TensorFlow NN models.}
\label{comb-latency-graph}
\end{figure*}

\section{Result}

\subsection{Model Latency Analysis}

 All models were evaluated using identical feature sets and window configurations to isolate architectural differences as the primary factor influencing latency and memory consumption.  In order to make the process reproducible and minimize measurement bias, latency profiling was performed entirely on-device rather than through external instrumentation.
 This methodology enables a fair comparison between classical machine-learning models and neural network approaches under identical hardware constraints.

We evaluated the real-time performance of each model by profiling inference latency directly on {\it WeBe Band}. To measure execution latency with minimal overhead, we implemented lightweight profiling hooks in the WeBe firmware using the CoreSight Data Watchpoint and Trace (DWT) unit. A 32-bit cycle counter was sampled immediately before and after each model call, and the cycle difference was translated to microseconds according to the 64~MHz system clock.

Table~\ref{tab:generic-metrics} summarizes the maximum inference latency for all evaluated models, each trained using the same set of 66 statistical features. Lightweight models such as \texttt{RF} and \texttt{PME} achieve inference latencies below 9~ms, whereas deeper neural network models exhibit higher latency due to increased depth and parameter count. These results demonstrate a strong dependence of the inference latency on the complexity of the model in resource-constrained environments.

\subsection{Memory Footprint and System-Level Performance Comparison}

{\it Piccolo AI} provides estimates of model memory usage and supports quantization of neural networks for microcontroller deployment, leveraging TensorFlow Lite for Microcontrollers (TFLM) as the underlying inference engine for quantized models. Table~\ref{tab:generic-metrics} also compares the use of Flash and SRAM across all evaluated models. Classical models such as \texttt{RF} and \texttt{PME} provide the best overall balance, combining low latency with modest memory requirements that comfortably fit within the constraints of MCU-class devices. 

Although the current implementation utilizes TFLM for neural network quantization and inference, the proposed open-source pipeline is modular and can accommodate alternative embedded inference backends, such as {\it CMSIS-NN} or other vendor-optimized libraries, with minimal architectural modification. This flexibility allows the workflow to adapt to different hardware ecosystems and evolving TinyML toolchains.

In contrast, TensorFlow-based neural network models require substantially more Flash memory (approximately 15--24~KB) to store network weights and exhibit higher SRAM usage due to intermediate activation buffers. Although these models offer greater representational capacity, their increased memory footprint and latency make them less suitable for highly resource-constrained embedded systems. However, in scenarios where higher classification accuracy or more complex decision boundaries are required, neural network models can serve as a robust alternative, provided that the additional computational and memory resources are available. Overall, the results highlight a clear trade-off between model complexity and deployability, with classical ML approaches offering more efficient real-time performance on wearable edge platforms.


Compared to prior TinyML studies that primarily evaluate model performance through offline benchmarks or simulation environments \cite{banbury2021benchmarkingtinymlsystemschallenges, warden2019tinyml}, our analysis emphasizes direct on-device profiling under realistic wearable constraints. Although absolute latency values depend on hardware characteristics, the observed trade-offs between classical machine-learning models and neural networks are consistent with previously reported trends. The integration of automated firmware deployment and real-time profiling distinguishes the proposed workflow from existing approaches by enabling rapid iteration and hardware-aware model selection within a unified development environment.

When analyzing inference latency as a function of sample index with a sliding window of 20 samples at a sampling rate of 25 Hz, all three models exhibit a periodic latency pattern, as shown in Fig.~\ref{comb-latency-graph}. Each model requires a window of 100 samples to produce an inference, and the sliding window advances by 20 samples at each step, resulting in a new inference approximately every 0.8 seconds while always operating on the most recent 4 seconds of data. The observed latency peaks correspond to these window updates, where a full inference pass is triggered. Importantly, the measured inference latency does not significantly interfere with continuous data acquisition; no samples are missed during execution, and data collection proceeds uninterrupted. The Random Forest (RF) and Pattern Matching Engine (PME) models exhibit relatively stable latency with limited variation across windows, whereas the neural network (NN) model shows higher latency and more pronounced peaks due to its increased computational complexity. Despite these differences, latency remains bounded and predictable across all models, confirming that sliding-window inference at 25 Hz can be executed reliably in real time on \emph{WeBe Band} without impacting data integrity.

\section{Summary and Conclusion}

In this paper, we presented a unified research framework that accelerates model development and automates firmware integration and on-device evaluation on \emph{WeBe Band}. Our system tightly integrates the {\it Piccolo AI} ecosystem with an automated deployment pipeline, streamlining the transition from model training to live execution on embedded hardware. Unlike prior TinyML efforts that primarily emphasize algorithmic innovation, this work focuses on system-level automation and deployability, enabling researchers to rapidly iterate on models without requiring deep expertise in embedded firmware development. The framework also supports hardware-aware compilation and OTA deployment, significantly reducing development time and technical overhead.

The profiling on the device reveals a clear trade-off between the complexity of the model and the feasibility of the system. Classical models such as Random Forest (RF) and Pattern Matching Engine (PME) achieve low latency and small memory footprints, making them well suited for real-time execution on resource-constrained platforms. In contrast, neural network (NN) models incur higher Flash and SRAM usage due to increased parameter storage and intermediate buffers. By abstracting embedded complexity through automated firmware integration and OTA deployment, the proposed framework enables rapid model iteration and informed model selection, and can be readily extended to additional sensing modalities.

This study intentionally prioritizes workflow automation and system-level evaluation over dataset scale or model generalization. The gesture dataset and limited user pool were selected to demonstrate deployability and real-time performance rather than clinical accuracy. Future work will expand evaluation to larger and more diverse datasets, incorporate additional physiological sensing modalities, and further investigate cross-platform deployment to validate scalability beyond the current prototype environment.

Beyond the immediate context of gesture recognition, the proposed workflow highlights a broader shift toward hardware-aware machine-learning research, where deployability and real-time performance become first-class design objectives. By enabling rapid experimentation directly on wearable hardware, the framework may facilitate interdisciplinary collaboration between clinicians, data scientists, and embedded engineers, accelerating the translation of TinyML research into practical healthcare applications.

As expected, the experimental results highlight that lightweight classical ML models provide an attractive balance between performance and resource efficiency on MCUs, while deeper neural networks must be carefully constrained to remain deployable. More importantly, the proposed workflow enables systematic exploration of these trade-offs directly on the target hardware, rather than relying solely on offline benchmarks.

\clearpage


\bibliographystyle{IEEEtran}
\balance
\bibliography{ref}

\end{document}